\documentclass[10pt,twocolumn,letterpaper]{article}

\usepackage{iccv}
\usepackage{times}
\usepackage{epsfig}
\usepackage{graphicx}
\usepackage{amsmath}
\usepackage{amssymb}
\usepackage{subfigure}
\usepackage{booktabs}
\usepackage{authblk}
\usepackage[table,xcdraw]{xcolor}

\usepackage[breaklinks=true,bookmarks=false]{hyperref}

\iccvfinalcopy 

\def\iccvPaperID{5} 

\ificcvfinal\fi

\begin{document}

\title{Residual Aligned: Gradient Optimization for Non-Negative Image Synthesis}

\makeatletter
\renewcommand\Authfont{\fontsize{11.5}{14.4}\selectfont}
\renewcommand\AB@affilsepx{\qquad \protect\Affilfont}
\makeatother
\author[1]{Flora Yu Shen}
\author[2]{Katie Luo}
\author[1,2]{Guandao Yang}
\author[1]{Harald Haraldsson}
\author[1,2]{Serge Belongie}
\affil[1]{Cornell Tech}
\affil[2]{Cornell University}
\renewcommand*{\Authsep}{ \ }%
\renewcommand*{\Authands}{ \ }%

\maketitle
\ificcvfinal\thispagestyle{empty}\fi


\begin{abstract}
In this work, we address an important problem of optical see through (OST) augmented reality: non-negative image synthesis.
Most of the image generation methods fail under this condition, since they assume full control over each pixel and cannot create darker pixels by adding light. 
In order to solve the non-negative image generation problem in AR image synthesis, prior works have attempted to utilize optical illusion to simulate human vision but fail to preserve lightness constancy well under situations such as high dynamic range. 
In our paper, we instead propose a method that is able to preserve lightness constancy at a local level, thus capturing high frequency details. Compared with existing work, our method shows strong performance in image-to-image translation tasks, particularly in scenarios such as large scale images, high resolution images, and high dynamic range image transfer.
 \end{abstract}

\section{Introduction}

Non-negative image generation problem \cite{StayPositive_2021} remains a difficult challenge in Augmented Reality (AR). Since the AR projector can only add light to the images, it is not possible to convert a lighter pixel to a darker pixel in an absolute way, let alone preserving the lightness constancy \cite{Adelson_24lightness}. Therefore, many works attempted to utilize human optical illusion \cite{StayPositive_2021, 9288440, MARINI19971655} in vision simulation. From AR projector relighting \cite{Huang_2021} to OST content optimization \cite{StayPositive_2021}, there have been many attempts to improve the overlaid content displayed on real world views; nevertheless, there still exists room for the visual quality to be improved. 

Prior works in field such as StayPositive formalized the problem as optimizing the residual image from state-of-art image generation methods. With an attempt to simulate human vision, their method assumes all parts of the image contributes to the visual output equally, and they solve it via a normalization over global maximum and minimum:
\begin{equation}
    N(x) = \frac{x - x_{min}}{x_{max} - x_{min}},
\end{equation}
where x is each pixel, $x_{min}$ and $x_{max}$ are the global maximum and minimum. However, the global normalization didn't preserve lightness constancy \cite{Adelson_24lightness} relative to surrounding neighborhoods, thus causing problems such as the input leaking through, and over-exposure artifacts in high dynamic range images. 

Inspired by lightness illusion \cite{10.1371/journal.pcbi.0030180}, color constancy illusions \cite{10.1007/3-540-63507-6_185, 10.1117/1.482707} and Poisson Image Editing \cite{Prez2003PoissonIE}, we designed a method that leverages gradient matching to preserve local features and blend a target image with the input image to realize image synthesis, with a soft constraint of non-negativity. This method leverages neighborhood region constraints and performs optimization over every single pixel through optimizing a gradients loss. After all, our method achieves strong results in ensuring lightness constancy on high dynamic range images, and preserving fine grained details on high resolution large scale images.

\section{Method}
\label{sec:method}
Because we observe humans are more prone to seeing \textit{changes} in darkness and brightness \cite{Land:71}, we are able to produce a better residual for human perception by focusing on lightness constancy at a local level.
In this section, we begin by discussing our two-part loss function inspired by prior work, then examine a local perceptual assumption --- gradient similarity --- and propose a computationally efficient method to incorporate it into the framework. Lastly we will go into some implementation details.
\subsection{Framework}
We formalize our problem as follow: we have input $I$, desired proposal $P$ generated from any state-of-the-art image-to-image translation method \cite{gan, gatys2016image, huang2017adain}, and we wish to optimize a residual $R_\theta$ that we overlay on the input.
By optimizing a two-part loss function, our method generates the final residual.

We assume that our final output image is a simple $\alpha$-blending between the input $I$ and the residual image $R_\theta$:
\begin{equation}
    O_\theta = \text{clip}\{\alpha I + (1 - \alpha) R_\theta\},
\end{equation}
where $\alpha$ is a property of the device under our optical combiner assumption, and $\alpha$ controls the effect of the input image.
To ensure physical feasibility, we clip the residual image into the desired range.

\begin{figure}
\includegraphics[width=\linewidth,trim={0.5cm 0 0 0},clip]{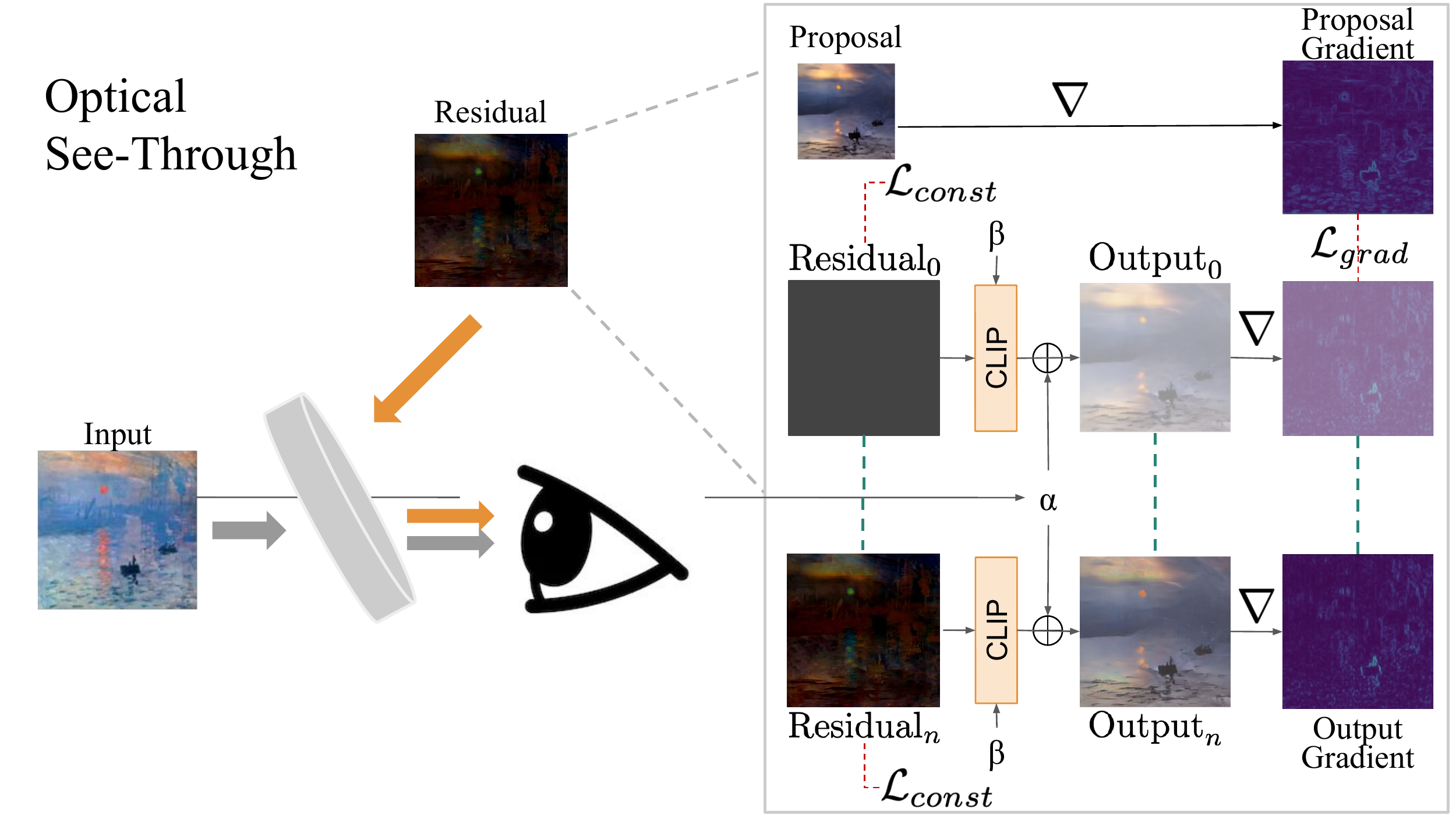}
\caption{Our model uses gradient to simulate perceptual similarity. We optimize every pixel in the residual image, by matching gradients with the target image.}\label{model}
\end{figure}

Our loss is comprised of the two desiderata for our image residual: the physical feasibility constraint loss of the residual $\mathcal{L}_{const}$, and the perceptual similarity loss between the target image and the final output $\mathcal{L}_{grad}$,
\begin{equation}
    \begin{aligned}
    \mathcal{L}(\boldsymbol{\theta}, I, P) =& ~\mathcal{L}_{const}(R_\theta, 0, 1 - \alpha) ~~+ \\
    &\mathcal{L}_{grad}(clip\{\alpha I + (1-\alpha)R_\theta\}, P).
    \end{aligned}
    \label{eq:final-ob}
\end{equation}

We employ a soft constraint loss to generate residuals that are as close to being physically feasible as possible.
\begin{equation}
\label{eq:constraint-loss}
\begin{aligned}
    \mathcal{L}_{const}(R, a, b) = \sum_{i,j}|\max(\min(R_{i,j}, b), a) - R_{i,j}|,
\end{aligned}
\end{equation}
where $R$ is the residual in our problem setting, and $a=0$ and $b=1$ are our minimum and maximum feasibility bounds. This way, we allow our model to make some mistakes by making the residual $R_\theta$ not \textit{completely} physically realizable at all pixels, as long as the error between the actual output and the target output is relatively small.

Our primary contribution lies in our perceptual similarity loss. Prior works \cite{StayPositive_2021} made the assumption that perceptual similarity occurred at a global level, and introduce a similarity loss that globally normalizes the images and then takes the Euclidean distance between them. This simulates the perceptual white-balancing. Furthermore, to simplify optimization, the StayPositive method used two global parameters on the entire image. 
This limits the range of images that can be expressed. In addition, human perceptual systems are not global across large regions, consisting of blind spots and sharper foveal vision. Our method utilizes the entire residual as individual optimizable parameters, thus increasing the flexibility of the final output image.

In order to match localized perceptual similarity, our method focuses on high-frequency detail losses by matching image \textit{gradients} in order to preserve lightness constancy at local features.

\subsection{Gradient Loss}
\label{sec:gradient_loss}

Inspired by seamless cloning and Poisson Image Editing \cite{Prez2003PoissonIE}, we decided to add constraint for pixels we want to fine-tune. 
This gives us the ability to match instance \textit{changes} in the proposal images, thus mimicking the way humans perceptual systems detect changes in patterns and edges.
However, unlike in Poisson Image Editing, we replace the hard constraint of value of the surrounded pixel, and use a soft constraint such that the modified pixel value should be non-negative. Also, in order to fine-tune the whole proposal images, we do not have a bounding area nor guidance field. Instead, we decided to use image gradients matching to ``blend" the input image and target image,

\begin{equation}
\label{eq:grad-loss}
\begin{aligned}
    \mathcal{L}_{grad}(O_\theta, P) = \Big\|\nabla_x O_\theta - \nabla_x P \Big\| + \Big\|\nabla_y O_\theta - \nabla_y P \Big\|,
\end{aligned}
\end{equation}
where $O_\theta = clip\{\alpha I + (1-\alpha)R_\theta\}$ is the output of our optical combiner.

By using gradient matching, we not only solve the problem with no target boundary, but we also allow the output to preserve local details and neighborhood features. 
For implementation, our gradient net uses the Sobel Operator \cite{sobelop} to calculate image gradients.
The Sobel Operator is traditionally used as an edge-detector algorithm, allowing us to match the high frequency details in our proposal image. 
Furthermore, it is a discrete differentiation operator, and gives us an efficient approximation of the gradient of the image intensity function.
Thus, we choose Sobel Operator because it emphasizes on high frequency details, is easy to implement, and is fast to compute.

\subsection{Implementation Details}
We start by initializing a zero-valued residual image, $R_\theta$, as the optimizable parameters. We clip the residual image into the feasible range, and minimize the total pixel value amount that it is out of range using the constraint loss from Equation~\ref{eq:constraint-loss}. Then, we combine the input $I$ and the residual image $R_\theta$ to obtain the output for first iteration. Lastly, we apply our localized similarity loss from Equation~
\ref{eq:grad-loss}.

For the gradient loss, we utilize the Sobel Operator with convolution kernel size of 3-by-3 to approximate the image intensity gradients.
We calculate image gradients on both horizontal $x$-direction and vertical $y$-direction. The gradient net processes image through RGB three channels for two directions. We take the Euclidean norm by stacking the 6 gradients together as a complete gradient for one image. This allows for efficient batch optimization.  

\section{Experiments}
\label{sec:experiments}
We validate our optimization method on the Day2Night \cite{Anokhin_2020_CVPR} images to test the robustness to bright and darkness changes; we also ran experiments on large scale, high resolution style transfer images \cite{Anokhin_2020_CVPR} to test the fine-grained detail performances.
We compared our method to heuristic baseline and the existing method StayPositive. However, since the StayPositive model optimized the residual image on two parameters (StayPositive-2parameters), while we optimized our residual image on all pixels, we extended the StayPositive method to an all pixels optimization (StayPositive-all pixels) to fairly compare the methods. Our method is able to out-perform both baselines, StayPositive-2parameters and StayPositive-all pixels. It demonstrates better performance in preserving dark and bright contrast in high dynamic range images, and finer details in large scale, high resolution images. We used Peak signal to noise ratio (PSNR), Structural Index Similarity (SSIM) score \cite{wang2004ssim}, and Learned Perceptual Image Patch Similarity (LPIPS) \cite{Zhang2018TheUE} to measure the performance. 

\begin{figure*}
    \centering
    \subfigure{\includegraphics[width=0.18\textwidth]{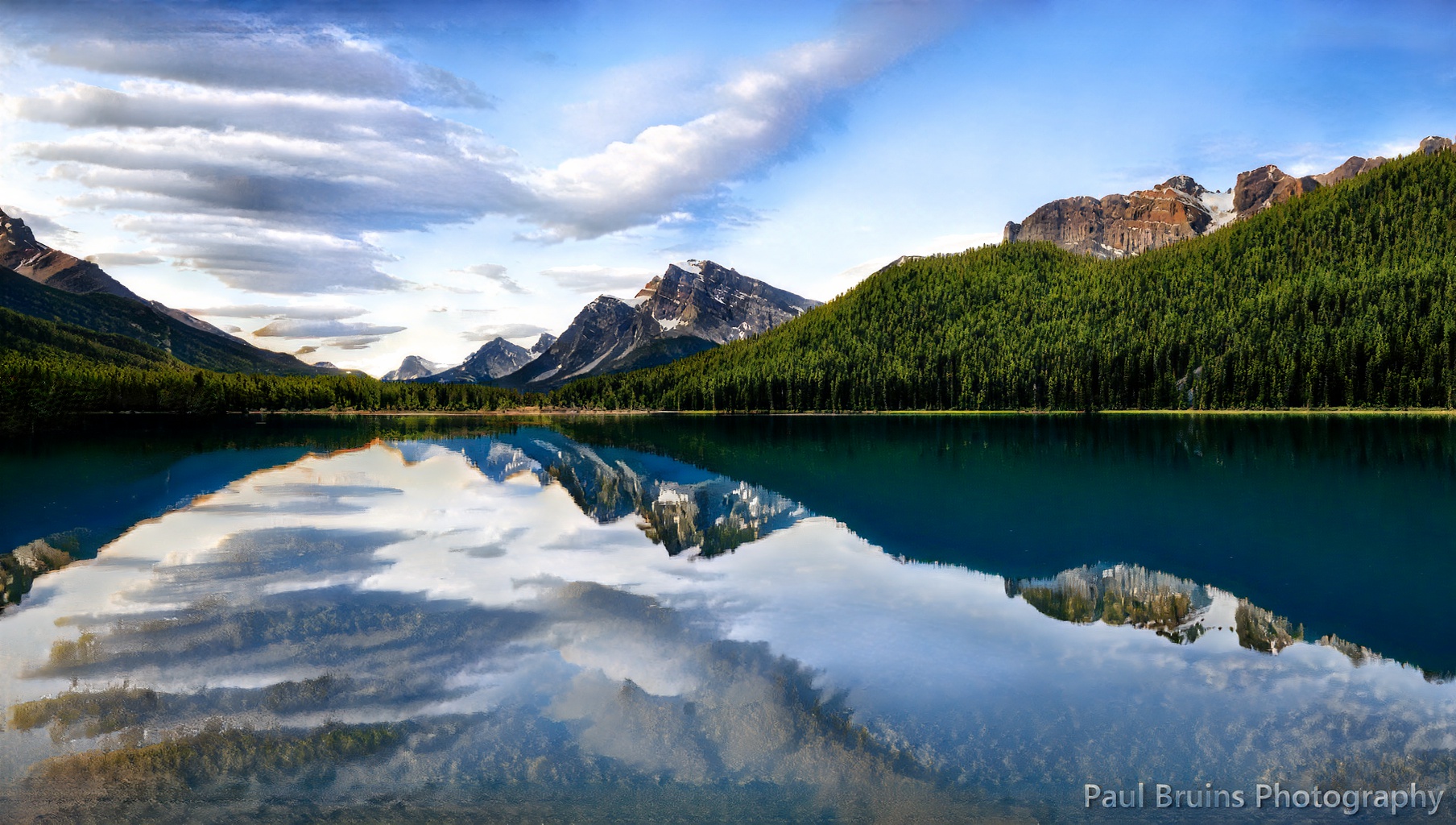}} 
    \subfigure{\includegraphics[width=0.18\textwidth]{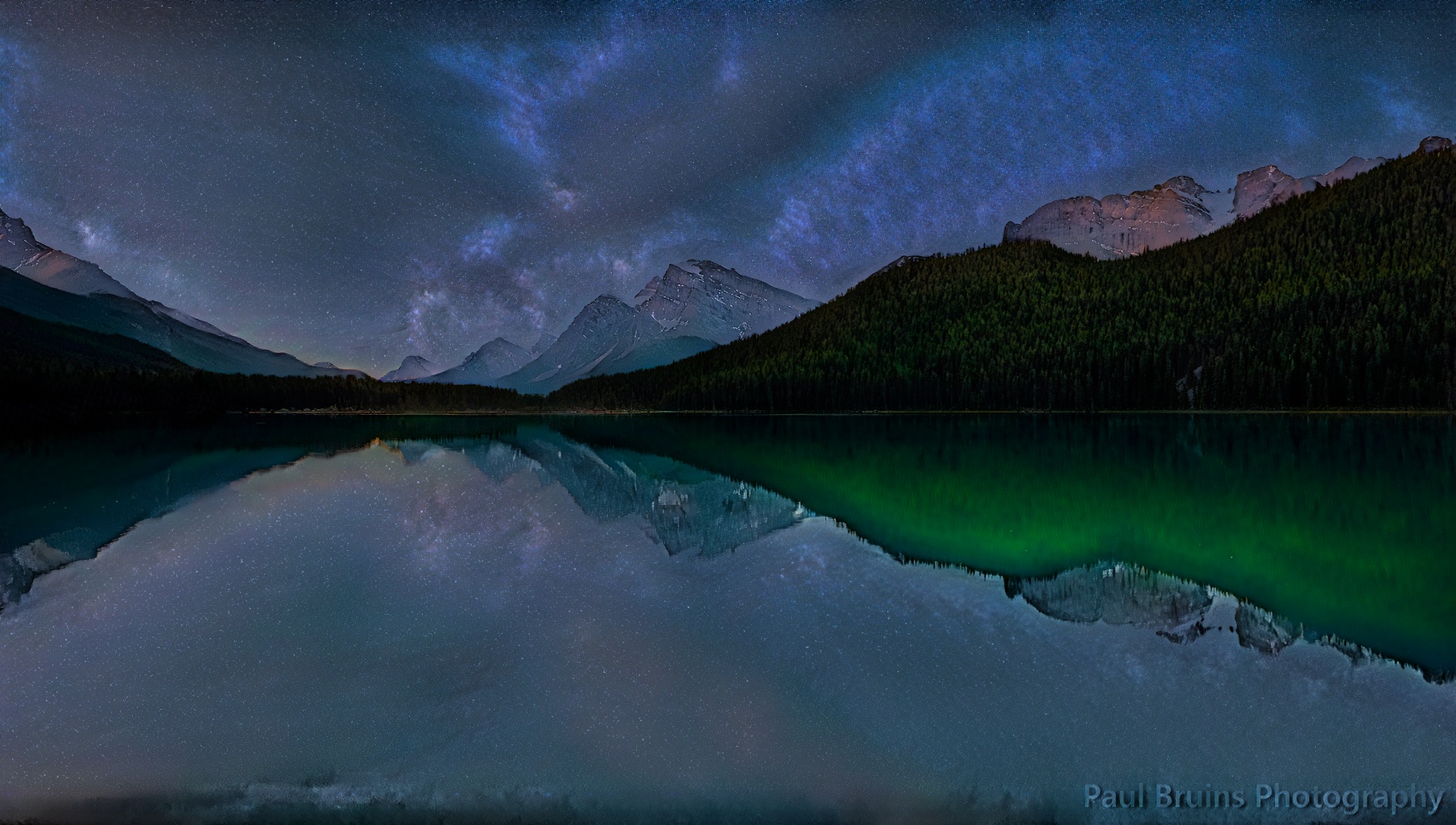}} 
    \subfigure{\includegraphics[width=0.18\textwidth]{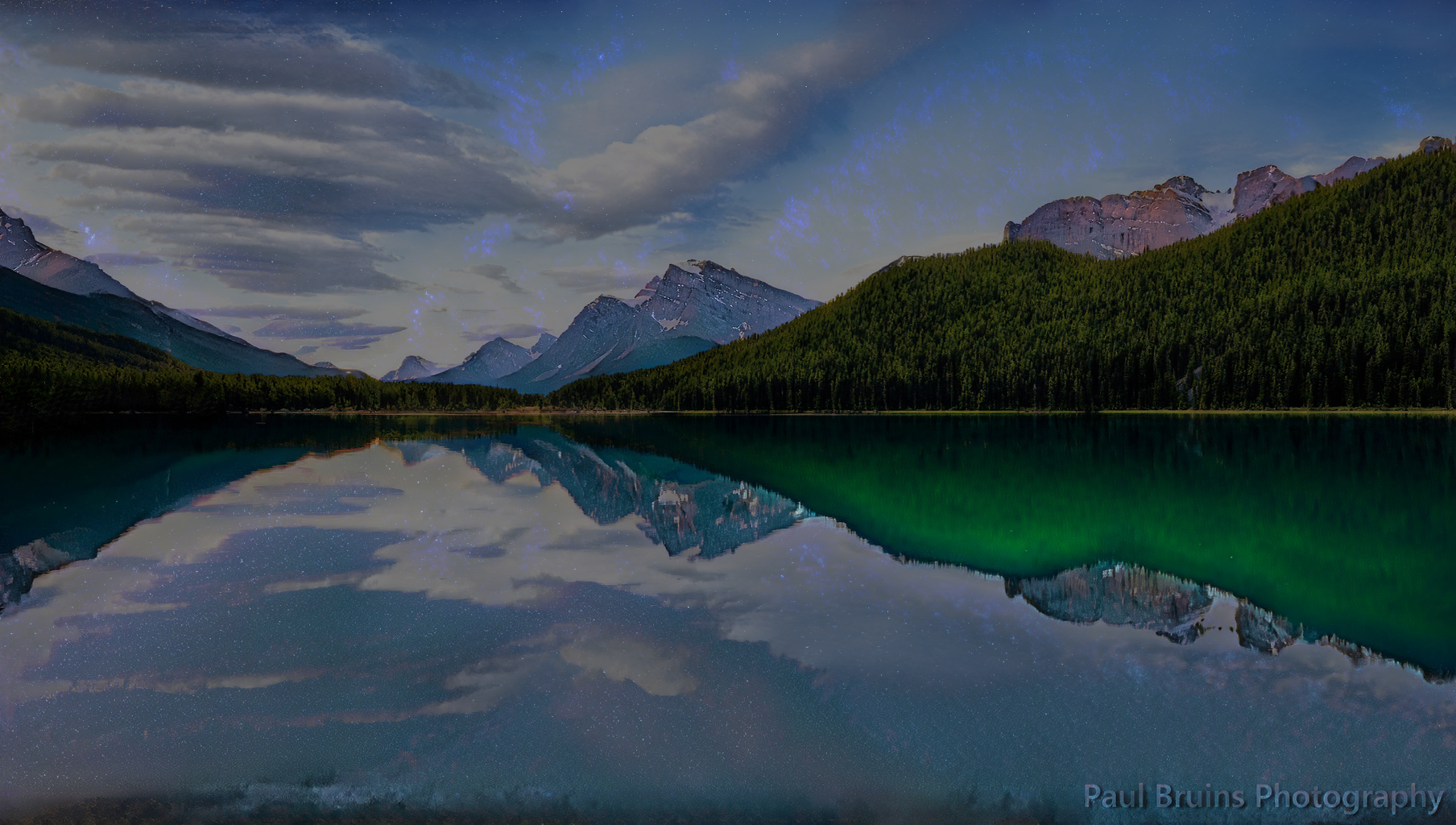}}
    \subfigure{\includegraphics[width=0.18\textwidth]{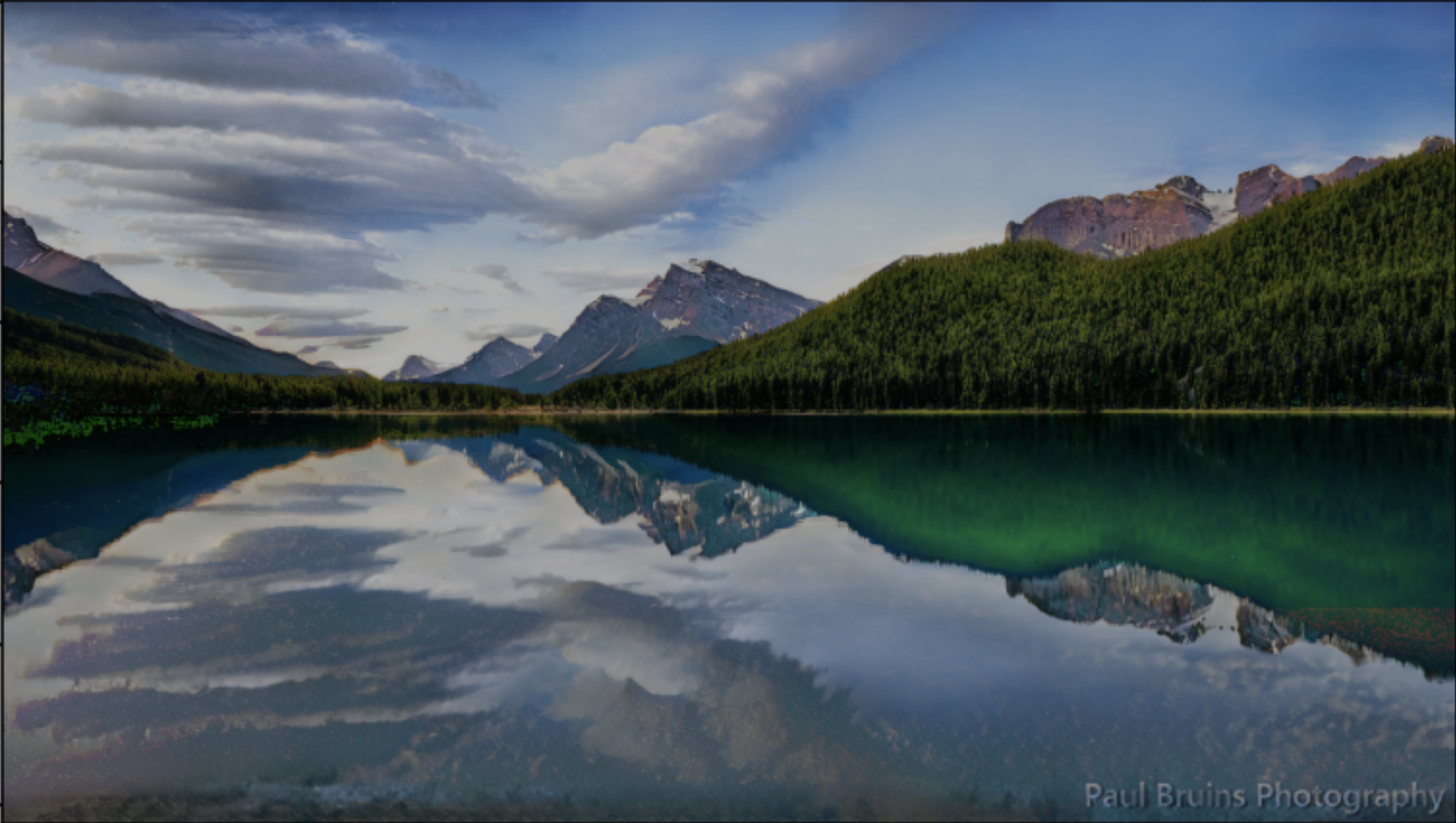}}
    \subfigure{\includegraphics[width=0.18\textwidth]{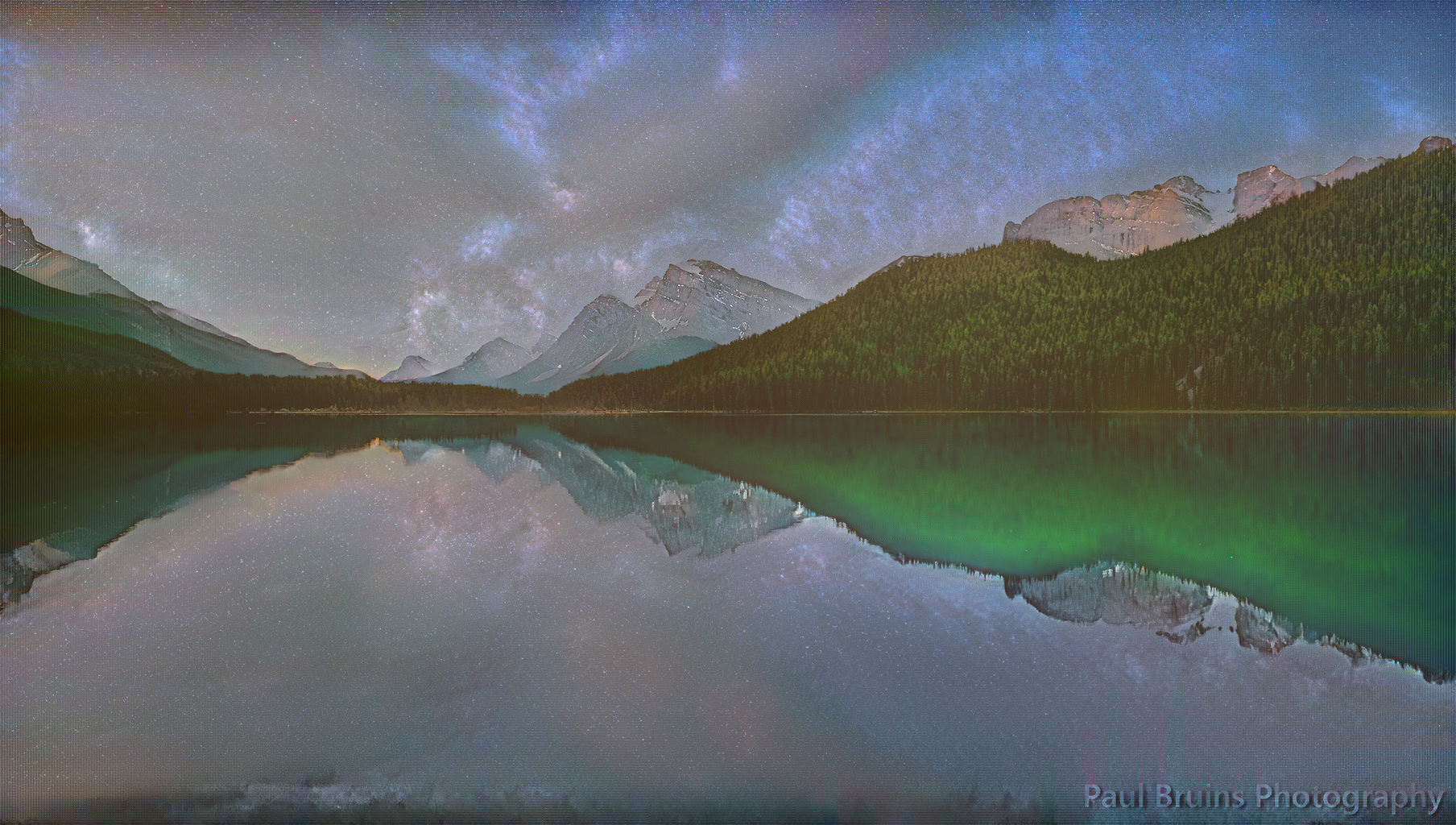}}
    
    \subfigure[Input]{\includegraphics[width=0.18\textwidth]{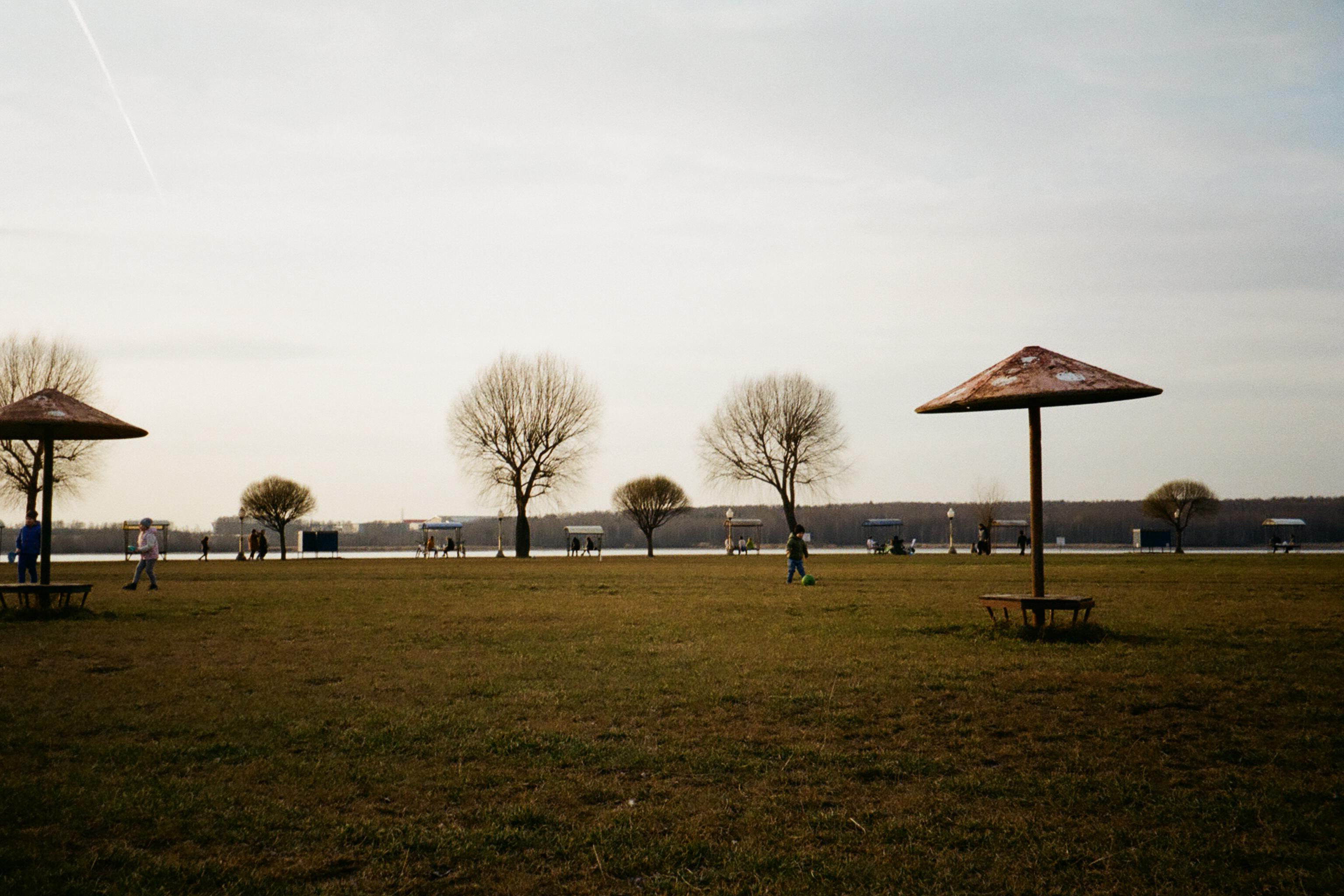}} 
    \subfigure[Target]{\includegraphics[width=0.18\textwidth]{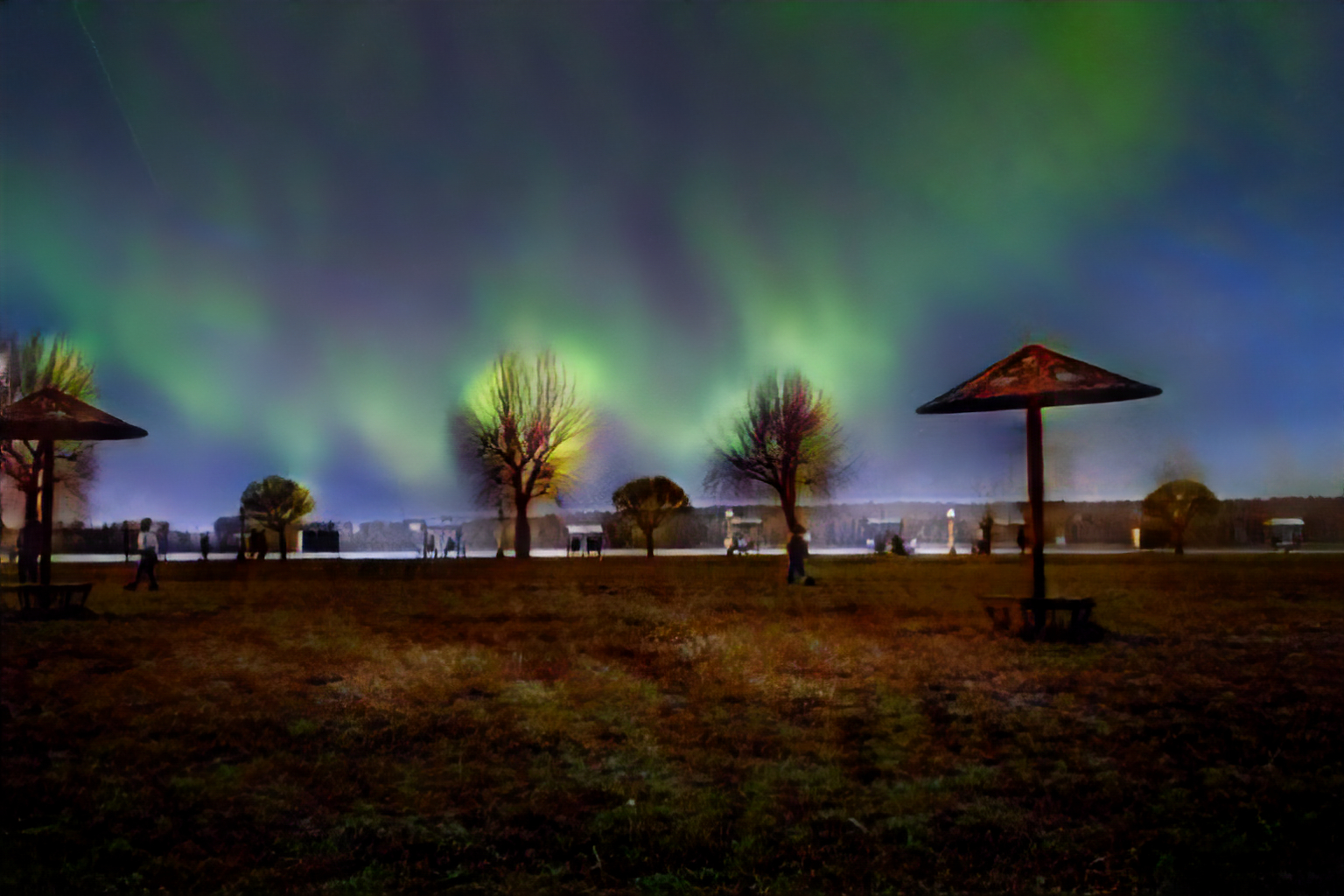}} 
    \subfigure[Heuristic]{\includegraphics[width=0.18\textwidth]{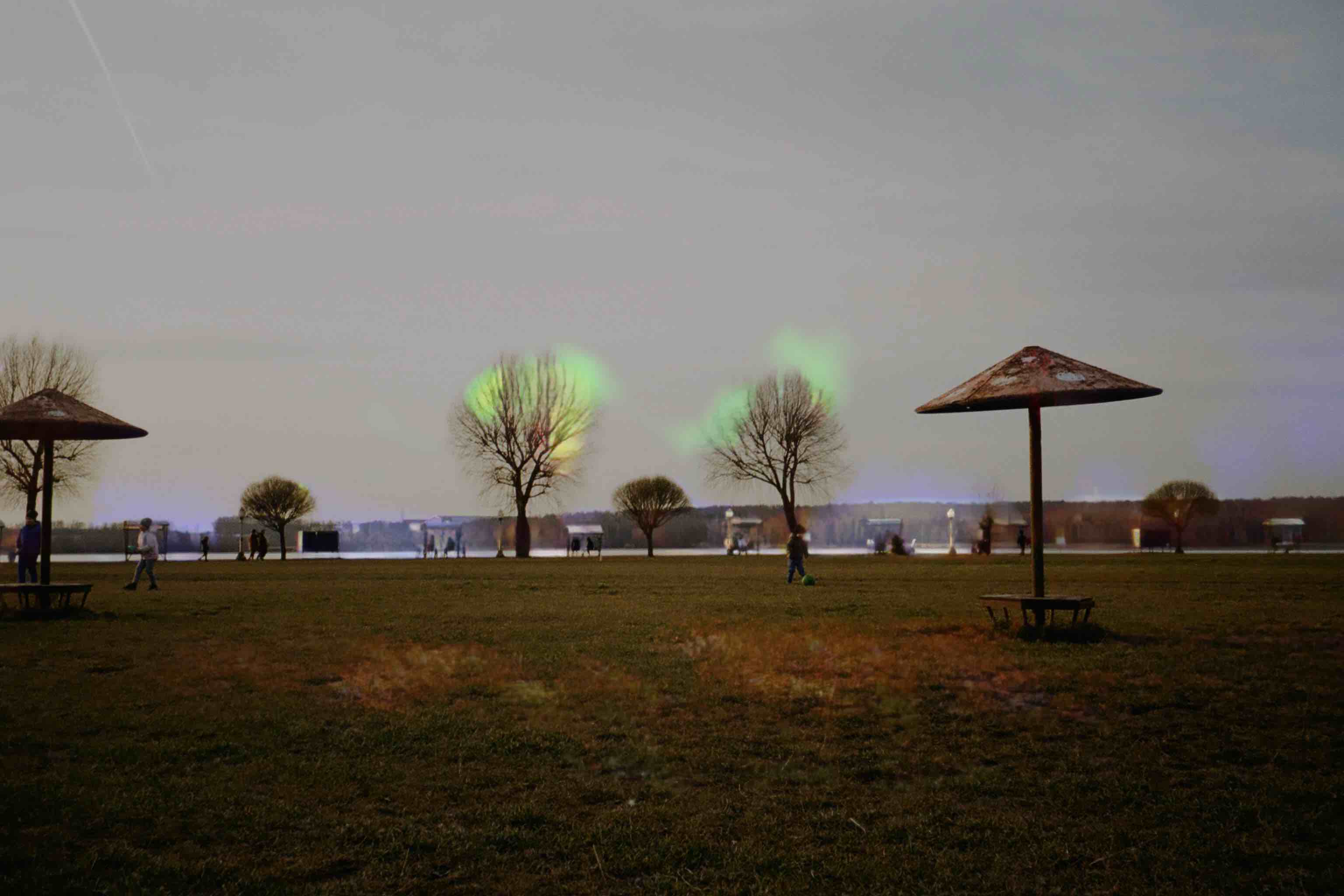}}
    \subfigure[StayPositive]{\includegraphics[width=0.18\textwidth]{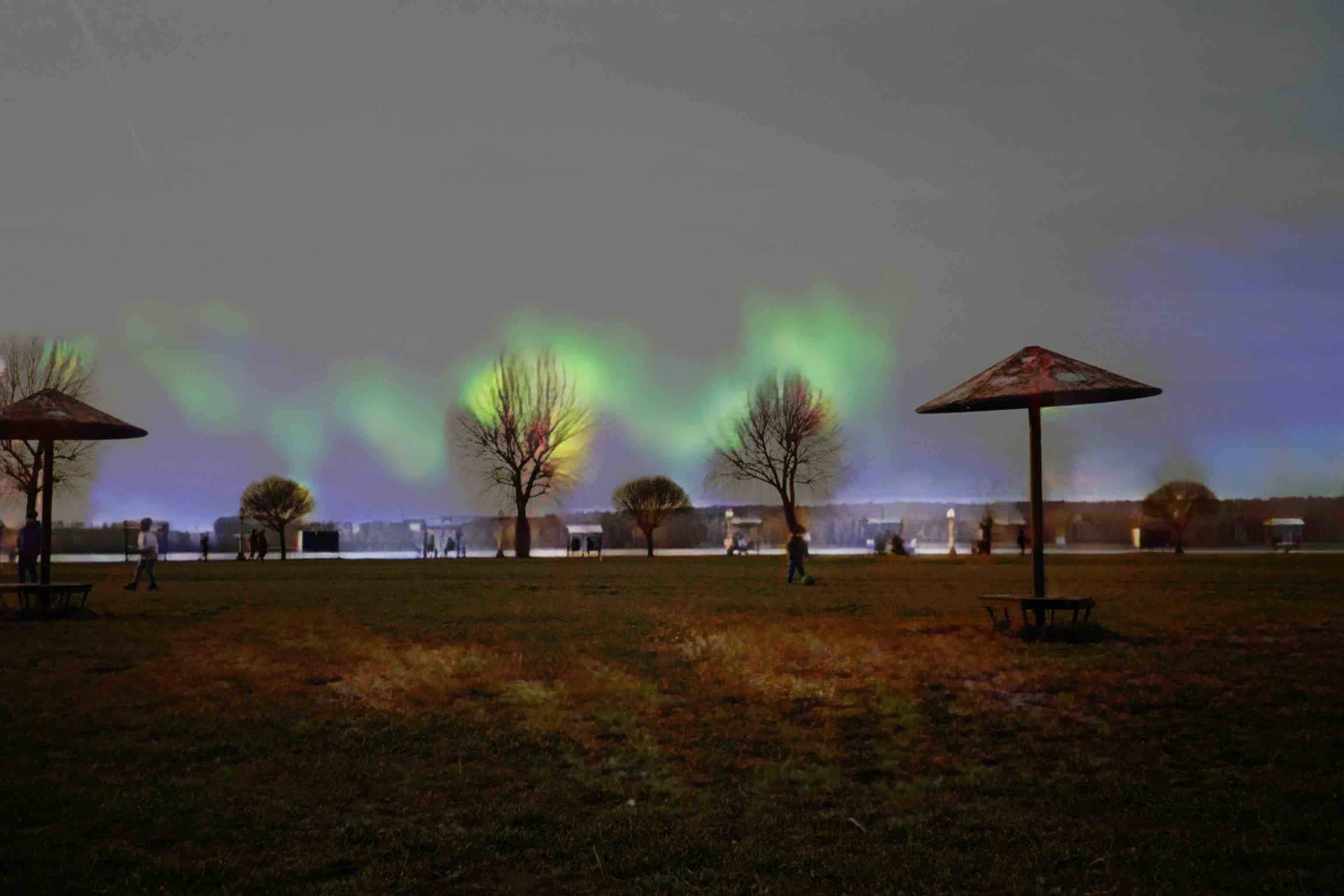}}
    \subfigure[Ours]{\includegraphics[width=0.18\textwidth]{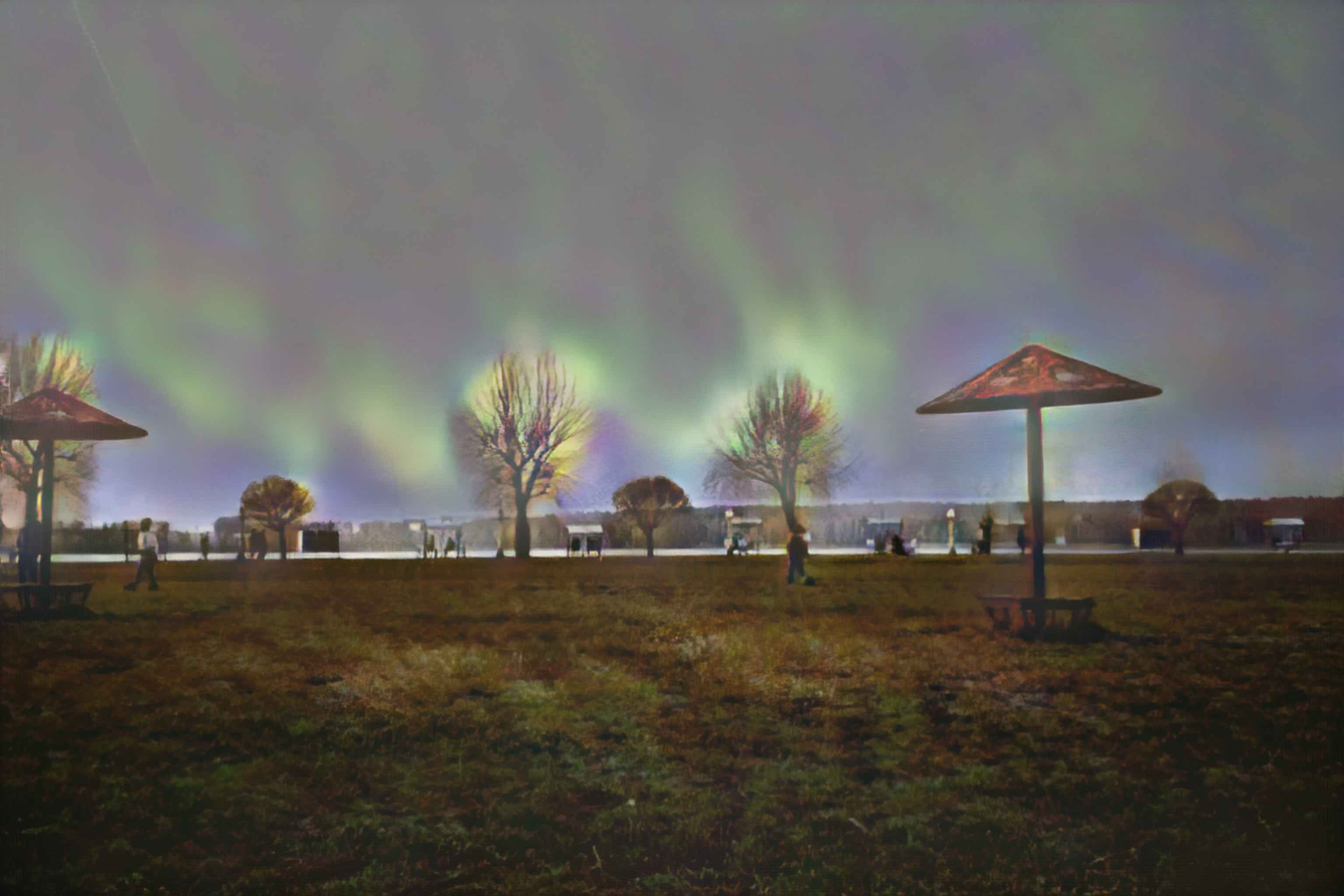}}

\caption{Our model is able to preserve the brightness contrast on high dynamic images with gradient optimization.}
\label{fig:examples}
\end{figure*}

\begin{figure*}
\centering
\includegraphics[width=\textwidth,trim={1.3cm 2cm 1.2cm 3cm},clip]{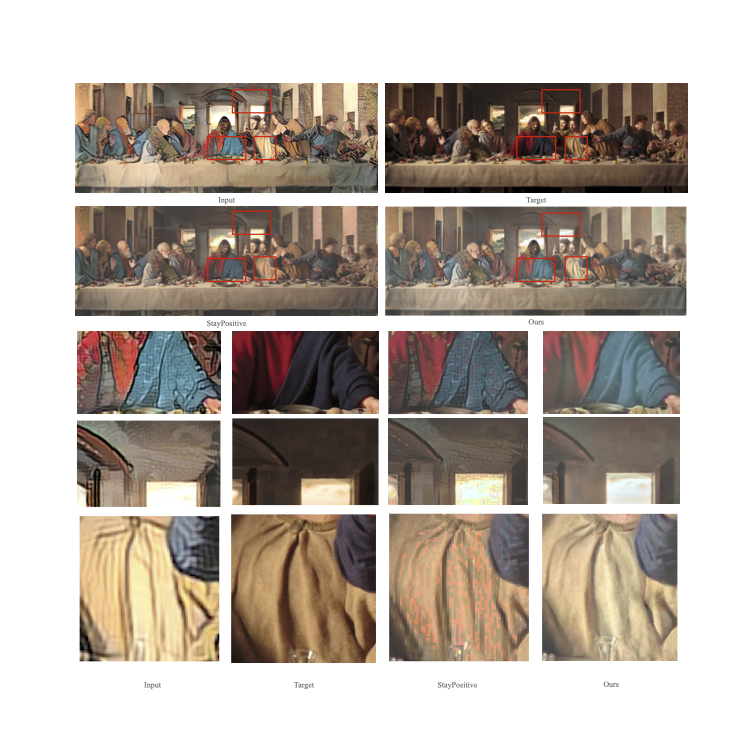}
\caption{Fine details comparison in large scale image (last supper). }
\label{fig:lastsupper}
\end{figure*}

\subsection{Performance on High Dynamic Range Images}
We believe by optimizing the gradient loss, we would be able to capture the relative pixels changes from the neighborhood, therefore capturing the contrast of darkness and brightness and project the gradient changes into pixel value movement. During experiments, we used the Day2Night image dataset to simulate the challenge of projecting light to bright (high pixel-intensity value) images. We compare our results with the heuristic baseline and the StayPositive-allpixels method. 
By matching gradients, the model takes advantages of human optical illusion of insensitivity to global lightness changes. Therefore even minor changes between dark and bright pixel values in very similar neighborhoods will be enlarged by human perceptual systems, regardless of the overall lightness. From \ref{fig:examples} first row, we can see that even though our method has an overall brighter effect compared to the target image, but human eyes will be deceived by the seemingly minor differences surrounded by neighborhood, and will perceive that the sky is dark and the scene is from nighttime. 
We compared the model performances using three metrics over averages on small patches of size $150\times150$. Our model has the highest PSNR score and lowest LPIPS score (perceptual loss), indicating that our method preserves the bright dark contrast well. 

\subsection{Performance on Large Scale Image Details}
Contrasting the StayPositive method of optimizing only two parameters for the residual images, we optimized each single pixels of the residual images. Therefore our method is able to preserve the local features and neighborhood gradient changes, exhibiting high-definition details especially in large scale images. Furthermore, by optimizing the gradient loss, our method is able to preserve the clear transition of edges, therefore demonstrating high frequency details especially in large images. During experiments, we used large scale, high resolution images such as Day2Night \ref{fig:examples} images and style transfer \ref{fig:lastsupper} images, which are over $4000\times1000$ pixels. We compare our results with the StayPositive-allpixels method. \par
From Figure \ref{fig:lastsupper}, we can see that our method shows finer and clearer details in the large scale image, with relative brightness and contrast preserved as well. The result is calculated by averaging metric on smaller patches because we want to compare similarity between local patches. Our method shows better results in all three metrics.

\begin{table}[]
\centering
\begin{tabular}{@{}lccc@{}}
\toprule
Methods            & PSNR($\uparrow$)  & SSIM($\uparrow$)   & LPIPS($\downarrow$)  \\ \midrule 
Heuristic Baseline & 14.9 & 0.542  & 0.641 \\
StayPositive       & 17.6 & \textbf{0.809} & 0.2704 \\
Our Method         & \textbf{18.6} & 0.6612 & \textbf{0.133}  \\ \bottomrule
\end{tabular}
\caption{This is the table for the high dynamic range images.}
\end{table}

\begin{table}[]
\centering
\begin{tabular}{@{}lccc@{}}
\toprule
Methods & PSNR($\uparrow$) & SSIM($\uparrow$) & LPIPS($\downarrow$) \\ \midrule
Heuristic Baseline  & 27.64       & 0.4662      & 0.453 \\
StayPositive    & 27.5        & \textbf{0.7075}      & 0.209 \\
Our Method  &\textbf{27.71}       & 0.699       & \textbf{0.100} \\ \bottomrule
\end{tabular}
\caption{This is the metrics table for the large scale last supper image. Metrics are computed on patches of 100$\times$100 pixels.}
\end{table}
\section{Conclusion}
In this work, we explored the non-negative image generation problem and proposed an effective non-deep learning image synthesis method, which can be combined with in-field image synthesis method and used in AR settings. It has the advantages in presenting fine details and preserving brightness contrast in large scale high resolution images and high dynamic range images. It achieve consistently better results than the baselines and existing methods. It can be used in interactive settings, to extend the visual enjoyment brought by augmented reality. 
However it also have some limitations. On small images, our method doesn't have clearly better performance. 
Under this setting, it may be more optimal to use a method that optimizes fewer parameters in a global way.
Furthermore, our framework assumes a highly simplistic optical combination framework. Future studies would need to look into how to more accurately model light combination from a projector with the real world under the OST setting.

{\small
\bibliographystyle{ieee_fullname}
\bibliography{main}
}

\end{document}